\documentclass[norunningheads]{styles/svproc}
\usepackage{url}
\usepackage{multicol}
\usepackage[bottom]{footmisc}

\usepackage{cite}
\usepackage{amsmath,amssymb,amsfonts}
\usepackage{algorithmic}
\usepackage{graphicx}
\usepackage{textcomp}
\usepackage{xcolor}
\usepackage{longtable}
\usepackage{caption}
\usepackage{booktabs}
\usepackage{hyperref}

\begin{document}
\mainmatter              
\title{Comparative Evaluation of Anomaly Detection
Methods for Fraud Detection in Online Credit Card
Payments}
\titlerunning{AD methods for Fraud Detection}  
%
\author{Hugo Thimonier\inst{1} \and Fabrice Popineau\inst{1}
\and Arpad Rimmel\inst{1} \and Bich-Li\^en Doan\inst{1} \and Fabrice Daniel\inst{2}}
\authorrunning{Hugo Thimonier et al.} 
%
\institute{Université Paris-Saclay, CNRS, CentraleSup\'elec,\\ Laboratoire Interdisciplinaire des Sciences du Num\'erique,\\ 91190, Gif-sur-Yvette, France \\
\email{name.surname@lisn.fr},\\
\and
LUSIS AI, Paris, France}
\maketitle              
\begin{abstract}
This study explores the application of anomaly detection (AD) methods in imbalanced learning tasks, focusing on fraud detection using real online credit card payment data. We assess the performance of several recent AD methods and compare their effectiveness against standard supervised learning methods. Offering evidence of distribution shift within our dataset, we analyze its impact on the tested models' performances.
Our findings reveal that LightGBM exhibits significantly superior performance across all evaluated metrics but suffers more from distribution shifts than AD methods. Furthermore, our investigation reveals that LightGBM also captures the majority of frauds detected by AD methods. This observation challenges the potential benefits of ensemble methods to combine supervised, and AD approaches to enhance performance. In summary, this research provides practical insights into the utility of these techniques in real-world scenarios, showing LightGBM's superiority in fraud detection while highlighting challenges related to distribution shifts.

\end{abstract}

\section{Introduction}
\label{sec:intro}

    Detecting fraudulent behaviors has emerged as a critical problem that has garnered significant attention from practitioners and scholars alike. In sectors such as banking, frauds incurred an estimated annual cost of \$28.58 billion in 2021, as highlighted by the Nilson Report 2021\footnote{https://nilsonreport.com/}. To address the challenge of identifying frauds within regular credit card payments, banks have increasingly turned to machine learning techniques, known for their effectiveness in many classification tasks, particularly with unstructured data.

    Two critical features of fraud detection pose challenges to constructing effective and accurate classifiers: highly imbalanced classes and distribution shift.
    
    Imbalanced datasets arise due to the significant disparity in the number of genuine transactions compared to fraudulent ones, making it difficult for traditional classification algorithms to generalize accurately. Highly imbalanced datasets are a specific subset of imbalanced datasets in which the positive class represents less than $1$\% of the samples; this situation is also referred to as rarity \cite{extreme_imabalance}. Moreover, distribution shift occurs as fraudsters constantly adapt their strategies, causing a discrepancy between the training and testing data distributions, thereby hampering the performance of machine learning models.

    Learning from imbalanced datasets is a critical topic with implications across various real-life applications. Extensive research has highlighted the consequences of imbalanced learning on canonical classifiers, revealing that most standard classifiers are ill-suited for imbalanced settings.
    For instance, the limitations of standard machine learning techniques when confronted with imbalanced datasets are examined comprehensively by \cite{Yanminsun2011}. This study also sheds light on the struggles faced by backpropagation algorithms in converging within imbalanced set-ups, as the dominant majority class can overwhelm the gradient vector used for weight updates in neural networks. In contrast, Gradient Boosted Decision Trees (GBDT) are often considered more resilient to imbalanced settings due to their focus on particularly challenging examples \cite{frery}, thus enabling them to prioritize the minority class more effectively. Highly imbalanced datasets are among the most challenging as research \cite{Japkowicz2002} has shown how learners display decreasing performance as imbalance becomes more severe.

    In addition to imbalanced datasets, distribution shift is another crucial challenge in detecting fraud. Standard machine learning techniques perform well when the training and testing datasets distributions are similar, if not identical. However, domain shift occurs when the distribution of the test dataset deviates from the original training distribution and thus hinders standard classifiers' performance.
    Fraud detection involves an iterative game between fraudsters and banks. Fraudsters continually strive to produce increasingly inconspicuous fraudulent behaviors, while banks aim to detect frauds as accurately as possible while avoiding false negatives. 
    This dynamic nature of fraud detection presents significant challenges for machine learning algorithms.
    
    These characteristics of fraud detection underscore the need for methodologies capable of effectively handling distribution shift and highly imbalanced datasets. In response to these challenges, researchers have proposed using anomaly detection (AD) methods, which promise to exhibit robustness in case of both distribution shift and extreme class imbalances. Specifically, AD involves the identification of anomalies within a dataset by delineating deviations from a predefined notion of normality. 
    Anomaly detection methods typically characterize the normal distribution solely based on normal samples during training. Consequently, AD has been regarded as particularly well-suited for imbalanced and extremely imbalanced settings. By design, AD methods do not experience performance deterioration when faced with highly skewed class distributions, as the training process solely requires normal samples. Moreover, assuming only fraudulent behaviors change over time, AD models should be more robust to distribution shift than standard supervised approaches. Indeed, if the normal distribution is well characterized, they should always be able to exclude new types of anomalies.

    In this work, we empirically investigated AD for fraud detection tasks, exploring their capabilities and limitations. By empirically evaluating various AD methods on a real-world dataset characterized by distribution shifts and extreme class imbalances, we aim to provide insights into the suitability and effectiveness of these techniques for addressing the challenges inherent to fraud detection. In addition to evaluating the performance of AD methods, we conduct a comparative analysis with Gradient Boosted Decision Trees (GBDT), the prevalent choice for machine learning tasks on tabular data \cite{grinsztajn2022why}, to gauge the added value of AD approaches. We rely on the LightGBM implementation \cite{ke2017lightgbm} and show that GBDT suffer significantly more from distribution shift than AD methods while displaying substantially better fraud detection performance than all tested AD methods. 
    Our paper is structured as follows: in the next section, we discuss works related to our application; in section \ref{sec:experiment_and_data} we discuss in detail the experiments conducted on our dataset; in section \ref{sec:results} we present the obtained results; in section \ref{sec:discussion} we discuss the results and finally in section \ref{sec:conclusion} we conclude.
    
\section{Related Works}
\label{sec:related-works}

    Anomaly detection encompasses two types of algorithms: supervised and unsupervised. In the case of supervised AD, one disposes of the label and indirectly uses it in the training process by building a training set solely composed of samples belonging to the \textit{normal} class\footnote{Throughout this paper, the term \textit{normal} relates to the notion of normality.}. Unsupervised AD methods involve situations where the label is unavailable, and anomalies must be directly identified within a dataset containing both \textit{normal} samples and anomalies.
    While supervised approaches can be used in situations where the imbalance is too severe for standard supervised approaches to work, unsupervised approaches are usually confined to applications that consist in removing samples that may hinder another models' performance on a particular task, e.g. mislabeled samples or outliers. 
    Since we dispose of labels in the context of fraud detection, we will focus on supervised anomaly detection.
    
    Supervised anomaly detection methods differ from standard supervised approaches because labels are only used indirectly. Indeed, standard supervised approaches consist in training a classifier using a dataset
    \begin{equation}
    \label{eq:d_train}
        \mathcal{D}_{train} =  \{(x_i,y_i): x_i \in \mathcal{X}, y_i \in \mathcal{Y} \}_{i=1}^n,
    \end{equation}
    using both sample features $x_i$ and labels $y_i$. Moreover, $\mathcal{D}_{train}$ contains samples from each class in $\mathcal{Y}$. On the contrary, supervised anomaly detection methods use the label to build a training set solely composed of a single class, referred to as the \textit{normal} class. In the case of binary classification, the \textit{normal} class is the majority class, e.g. the legit payments in the case of fraud detection, and the training set can then be constructed as
    \begin{equation}
        \label{eq:d_train_ad}
        \mathcal{D}_{train}^{AD} = \{x_i: y_i = 0\}_{i=1}^n.
    \end{equation}
    In this anomaly detection framework, the overall goal is to characterize the normal distribution, $p(x\mid y=0)$. In inference, this characterization is used to determine whether a sample belongs to the normal distribution or should be seen as an anomaly.
    
    As a field of research, anomaly detection can be divided into several non-exhaustive categories: one-class classification (OCC), reconstruction-based methods and self-supervised methods.

    \paragraph{One-Class Classification}
        In contrast to traditional machine learning classification problems, one-class classification (OCC) approaches aim to identify samples that do not belong to a specific class by characterizing the distribution of that class. These discriminative models learn a decision boundary using only samples from the designated \textit{normal} class, thereby circumventing the direct estimation of the class distribution.
        During the inference phase, samples are classified as either belonging to the \textit{normal} class or not, without making any assumptions about the \textit{anomaly} class. 
        One-class support vector machines (OCSVM) \cite{ocsvm} and support vector data description (SVDD) \cite{svdd} are popular OCC methods that rely on kernels to map the data space to a Hilbert space, where a decision boundary is learned. 
        Recently, \cite{ruffDeepOneClassClassification, ruffDeepSemiSupervisedAnomaly2020} have introduced extensions to OCC methods that incorporate deep neural networks to alleviate the computational complexity associated with kernels.
        Other OCC tree-based approaches can be found in the OCC such as isolation forest (IForest) \cite{isolationforest}, extended isolation forest \cite{ief}, Robust Random Cut Forest (RRCF) \cite{Guha2016} and PIDForest \cite{Gopalan2019PIDForestAD}.
        Other methods have relied on sample-sample dependencies to identify anomalies such as TracInAD \cite{thimonier2022} relying on influence measures or approaches based on k-nearest neighbors (KNN). In the latter, anomalies are identified by measuring the distance of each sample to its k-nearest neighbors \cite{angiulli2002fast, ramaswamy2000efficient}: higher distance indicating abnormality. 
        
    \paragraph{Reconstruction-based methods}
    \label{subsec:reconstruction-based-methods}
    
        Reconstruction-based anomaly detection methods rely on the assumption that different distributions generate normal samples and anomalies.
        Consequently, training a model to reconstruct samples from the \textit{normal} distribution aims to achieve low reconstruction error for any sample belonging to this distribution. Conversely, anomalies that are believed to stem from a distinct distribution should exhibit significantly higher reconstruction errors.

        One of the most prevalent shallow reconstruction-based anomaly detection methods employ Principal Component Analysis (PCA) or Bayesian PCA \cite{Dutta2007, Scholkopf2007}.
        Autoencoders \cite{Finke2021},  regularized autoencoders like Variational Autoencoders (VAEs) \cite{pol2019} and memory-augmented deep autoencoders \cite{Gong2019MemorizingNT} have also been leveraged for anomaly detection. Recently, \cite{Kim2020RaPP} proposed a novel methodology for anomaly detection using autoencoders that incorporates the hidden representations of the original and reconstructed samples. Instead of solely comparing the reconstructed sample and the original sample, the authors suggest comparing the hidden representations of both samples by passing the reconstructed sample through the autoencoder. In addition, recent approaches have explored attention-based architectures for reconstructing masked features of samples, as exemplified by NPT-AD \cite{thimonier2023individual}. Other related methods do not directly compute a reconstruction error and only focus on estimating either the entire \textit{normal} distribution such as ECOD \cite{ecod} or local \textit{normal} distributions as proposed in local outlier factor (LOF) \cite{lof}.
    \paragraph{Self-supervised methods}
    The literature also features self-supervised approaches employing pretext tasks for anomaly detection \cite{bergmanClassificationBasedAnomalyDetection2020, qiuNeuralTransformationLearning2021a, TackMJS20}. In GOAD \cite{bergmanClassificationBasedAnomalyDetection2020}, several affine transformations are applied to each sample in the training set, while a classifier is trained to predict the specific transformation applied to a transformed sample. During testing, since the classifier was exclusively trained on \textit{normal} samples, it is expected to struggle in correctly predicting the transformation for anomaly samples. Similarly, \cite{qiuNeuralTransformationLearning2021a} propose NeuTraL-AD, a contrastive framework in which they transform samples using neural mappings instead of affine transformations. The objective is to learn transformations that maintain similarities in a semantic space between transformed samples and their untransformed counterparts while different transformations are easily distinguishable. In inference, the contrastive loss utilized to optimize the parameters serves as the anomaly score.
    More recently, \cite{shenkar2022anomaly} introduced a self-supervised methodology for anomaly detection that maximizes the mutual information among the elements of a sample's features using contrastive learning. By maximizing the mutual information, the method effectively captures the underlying structure of normal samples and identifies deviations indicative of anomalies.

    \paragraph{Supervised classification on tabular data}
    Although deep-learning models have become ubiquitous for a broad range of tasks involving natural language processing (NLP) and computer vision (CV), applying these models to tabular data remains very challenging. 
    Some recent methods \cite{kadra2021well, gorishniy2021revisiting, saint2021} have shown promising results when applying deep learning models tailored for tabular data. However, in recent work \cite{grinsztajn2022why, gorishniy2023tabr}, authors discuss how neural networks tend to struggle with this data type in comparison with other methods based on gradient-boosted decision trees (GBDT). In most scenarios, approaches such as XGBoost \cite{xgboost} or LightGBM \cite{ke2017lightgbm} have been shown to surpass deep learning algorithms. This type of approach remains the go-to method for practitioners due to its strong classification performance and its simplicity to train in comparison with deep methods. Moreover, GBDT models such as LightGBM and XGBoost are often considered particularly suited for imbalanced and extremely imbalanced set-ups since these models focus on particularly hard-to-classify samples \cite{frery}, generally the minority class, and thus offer strong performance in comparison to other standard machine learning models.

\section{Experiments and Datasets}
\label{sec:experiment_and_data}

    \subsection{Dataset}
    \label{subsec:dataset}

    We dispose of a labeled dataset of online credit card payments made available by a major French bank. Our dataset contains 145 features describing raw characteristics of payments (e.g. amount, currency) as well as features that we computed, such as rolling sums or rolling means.
    Our dataset contains 480 million online transactions from the first day of 2018 until the last day of 2021. The dataset is comprised of two independent datasets merged, one from 2018 to 2019 and the other from 2020 to 2021. This dataset displays the characteristic of highly imbalanced classes\footnote{Due to confidentiality, we cannot discuss the exact proportion of frauds within our dataset.} discussed in section \ref{sec:intro} since the proportion of legit payments vastly outnumbers the proportion of frauds. 
    We remove cards with less than 50 payments, cards for which the proportion of frauds exceeds $50$\%, and cards with too few payments since they would not have derived features (e.g. rolling means) with meaningful values and would risk hindering the models' performance. Similarly, we also omit cards with too many frauds in their payment history since they would also be problematic as they might pollute the fraud distribution. Overall, this preprocessing reduces the dataset to $192$ million payments. Most methods we wish to test would be intractable with such dataset size and require further dimension reduction. Thus, we restrict our analysis to two countries with $3$ million payments ($1.5$\% of total dataset size) and $20$ million payments ($10.3$\% of total dataset size), respectively. Moreover, restricting our analysis to one country at a time should help models learn since payment distributions likely differ between countries.

\subsubsection{Distribution shift}

\begin{figure*}[tb]
  \centering
  \begin{minipage}[b]{0.49\textwidth}
    \centering
    \caption*{Country A}
    \includegraphics[width=\textwidth]{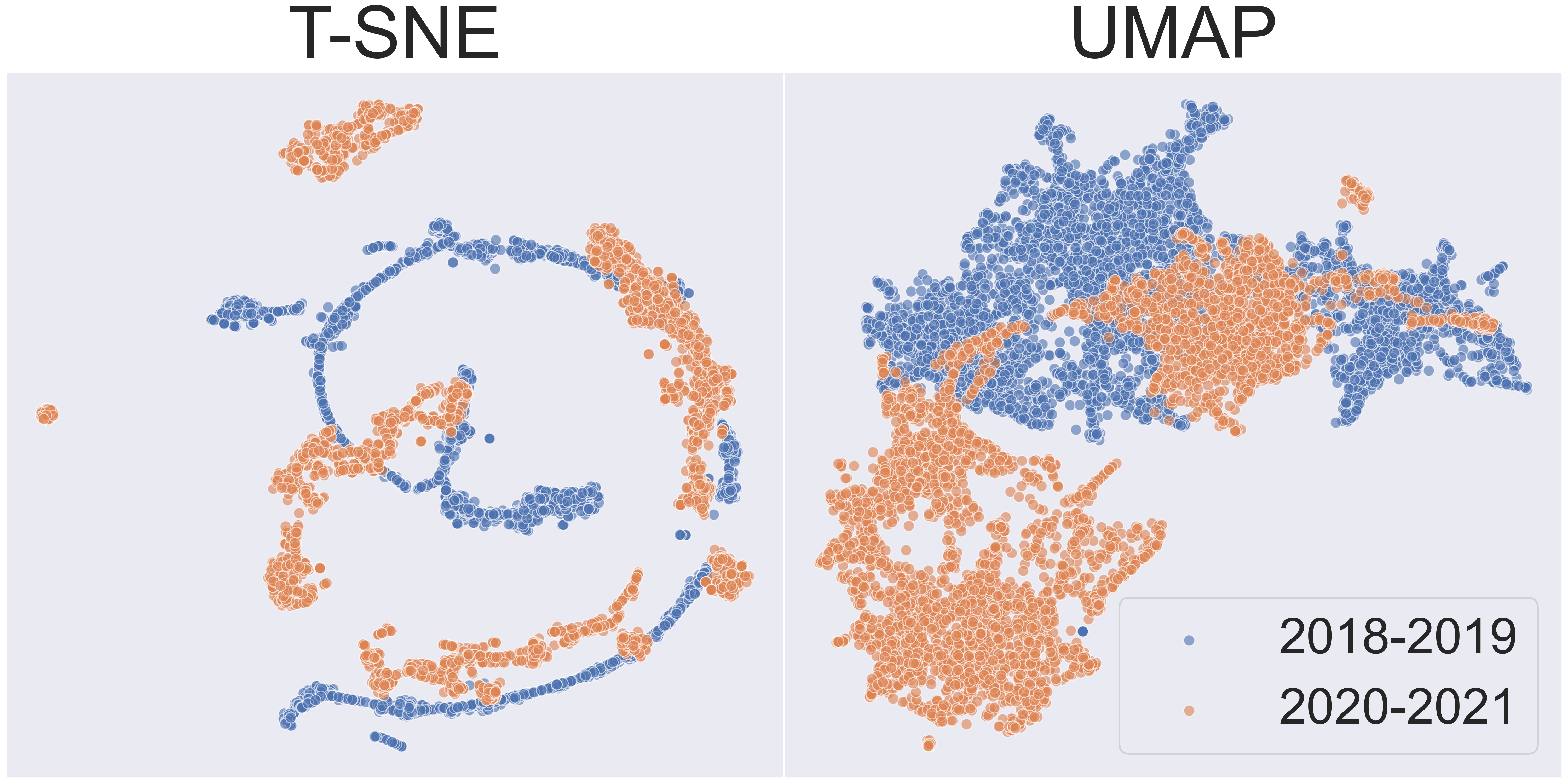}
    \vspace{10pt}
    \begin{tabular}{ccc} \toprule
        OTDD & $2018$ & $2020$ \\ \midrule
        $2018$ & $56.5$ & $\mathbf{64.4}$ \\
        $2020$ & $\mathbf{61.9}$ & $53.1$ \\ \bottomrule
    \end{tabular} 
  \end{minipage}
  \hfill
  \begin{minipage}[b]{0.49\textwidth}
    \centering
    \caption*{Country B}
    \includegraphics[width=\textwidth]{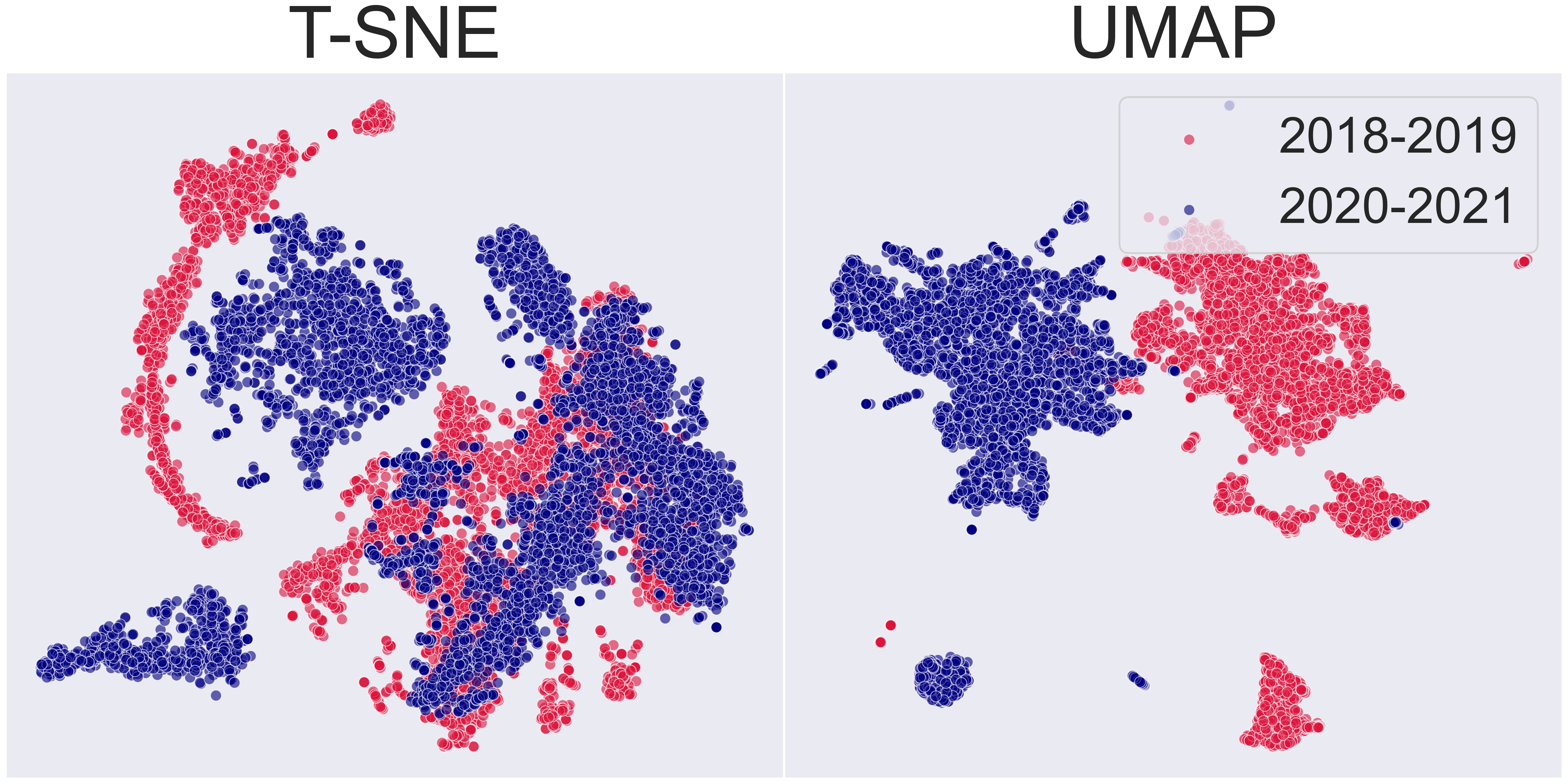}
    \vspace{10pt}
      \begin{tabular}{ccc} \toprule
          OTDD & $2018$ & $2020$ \\ \midrule
          $2018$ & $54.7$ & $\mathbf{69.2}$ \\
          $2020$ & $\mathbf{73.6}$ & $56.9$ \\ \bottomrule
      \end{tabular}
  \end{minipage}
  \caption{T-sne \cite{vanDerMaaten2008} and UMAP \cite{mcinnes2020umap} bi-dimensional representation of payments in countries A and B for each period. These graphs give evidence of a distribution shift for both countries' payment behaviors between 2018 and 2020 since we observe very few sample overlays. Tables give the Optimal Transport Dataset Distance \cite{Alvarez-MelisF20} between subsamples for each country. We observe a much higher distance between subsamples from the same country between different periods than for the same period}
  \label{fig:dist_shift}
\end{figure*}

    We argue that our datasets undergo a distribution shift and that the distributions of legit and fraudulent payments differ between 2018 and the end of 2021. For instance, as supported by \cite{jtaer16060125} Covid-19 has caused online payment behaviors to change over time drastically. To further support our statement, we display in figure \ref{fig:dist_shift} the t-sne \cite{vanDerMaaten2008} and UMAP \cite{mcinnes2020umap} representations for years 2018-2019 and 2020-2021 for both countries. We observe significant dissimilarities between datasets. For country A, we observe an entire subsample of payments made in 2020-2021 at the bottom left of the UMAP graph, which does not exist for the 2018-2019 period. Similarly, for the t-sne representation of country A, we observe a similar pattern with few data sample overlays between periods.
    Moreover, for country B, we also observe a very scarce overlay on the graph between periods, especially for the UMAP representation. To further investigate whether distribution shift is present in our datasets, we rely on the Optimal Transport Dataset Distance method (OTDD) \cite{Alvarez-MelisF20} to measure the distance between datasets from each period. This method relies on optimal transport, which measures the distance between distributions. For each country, we created two subsamples of 5000 observations for each period and compared the distance between the subsample of the same period and between periods. Results of this analysis are shown in the tables in figure \ref{fig:dist_shift} and indicate that the distance between the dataset increases across periods. This is especially true for country B. 

    \subsubsection{Data splits and preprocessing}
    
    For a fair comparison between supervised approaches and anomaly detection methods, we split the 2018 datasets of each country into two separate datasets constituted of legit payments and frauds. We take a training set representing 75\% of the 2018 dataset for each country and use a test set the 25\% remaining. We include the frauds in the training set for LightGBM, while for anomaly detection approaches, the frauds are excluded from the training set. The 2020 dataset of each country serves entirely as test sets.
    Overall, the considered dataset used for every tested model contains features describing characteristics of the payment (e.g. amount of the transaction, the currency used, duration since the last transaction.), among which eight are categorical. All eight categorical features are encoded using Catboost encoding following \cite{catboostencoding}. Continuous features are scaled to be in $(0,1)$ through standard normalizing by removing the mean and reducing to unit variance.

    \subsection{Experimental Settings}
    \label{subsec:expe-settings}

        In order to evaluate the suitability of anomaly detection methods for fraud detection, we conduct a comprehensive analysis using state-of-the-art approaches on our dataset. We investigate both deep learning-based techniques designed for tabular data, such as the self-supervised approaches of GOAD \cite{bergmanClassificationBasedAnomalyDetection2020}, NeuTraL-AD \cite{qiuNeuralTransformationLearning2021a} the contrastive approach proposed in \cite{shenkar2022anomaly}, and the reconstruction-based approach of NPT-AD \cite{thimonier2023individual}. Additionally, we include non-deep learning methods, namely Isolation Forest \cite{isolationforest}, ECOD \cite{ecod}, COPOD \cite{copod2020}, and the KNN AD approach \cite{angiulli2002fast, ramaswamy2000efficient}, as they have demonstrated remarkable performance on various tabular datasets. As a baseline, we employ LightGBM \cite{ke2017lightgbm}, a well-established supervised classification method, to assess the added value of anomaly detection compared to standard supervised techniques in this context.

        To effectively compare the performance of various models in detecting fraud, we employ three commonly used metrics from the anomaly detection literature and the banking industry for real-life model evaluation.
        The anomaly detection literature has widely adopted the F1-score and the AUROC as evaluation metrics. While the AUROC is suitable for balanced class distributions, it may not fully account for class proportions, which is crucial in assessing performance in imbalanced set-ups. We include the Area Under the Precision-Recall Curve (AUPRC) to address this limitation, which is better suited for imbalanced datasets. In support of this choice, \cite{auprcvsauroc} has demonstrated that a model dominates in the ROC space if and only if it also dominates in the PR space.
        
        For the F1-Score, whose value depends on a threshold, we adhere to the practices of the anomaly detection literature by selecting the threshold for the anomaly score that predicts an equal number of fraud cases as those present in the dataset. This approach ensures a fair and consistent evaluation across all models.

    \section{Results}
    \label{sec:results}

    \subsection{Models Hyperparameters}
    \label{subsec:hyperparam}

        We implemented the non-deep models using the PyOD library \cite{zhao2019pyod} with default hyperparameter values and LightGBM \cite{ke2017lightgbm} also with default parameters. For the deep learning approaches, we adopted the hyperparameters suggested in the original papers for the dataset that most resembled our datasets. Specifically, we set the hyperparameters to those used for the KDD dataset for NeuTraL-AD \cite{qiuNeuralTransformationLearning2021a}, GOAD \cite{bergmanClassificationBasedAnomalyDetection2020}, and NPT-AD \cite{thimonier2023individual} using their official implementations available on GitHub. Regarding the approach proposed by \cite{shenkar2022anomaly}, we kept the parameters at their default values as specified in their implementation.
        Deep models were trained on 4 Nvidia GPUs V100 16Go/32Go, while non-deep models were trained on 2 Intel Cascade Lake 6248 processors (20 cores at 2.5 GHz), thus 40 cores.
    
    \subsection{Fraud Detection Performances}

        \begin{table*}[h!]
          \centering
            \caption{Performance metrics of AD models in comparison with LightGBM \cite{ke2017lightgbm}. Results are averaged over 10 runs for 10 different splits of the data, the standard deviation are displayed below the metrics. We report the F1-Score in terms of percentage. The highest metric over all models is highlighted in bold, while the highest metrics among the AD method are underlined. We perform $5$\% t-test between the highest metrics to measure whether they are statistically different}
          \resizebox{\textwidth}{!}{
          \begin{tabular}{c cccccc|cccccc}
          \toprule
          & \multicolumn{6}{c}{\textbf{Country A}} & \multicolumn{6}{c}{\textbf{Country B}} \\
          \midrule
          Model &
          \multicolumn{2}{c}{F1 $(\uparrow)$}
          & \multicolumn{2}{c}{AUROC $(\uparrow)$}
          & \multicolumn{2}{c}{AUPRC $(\uparrow)$} 
          &  \multicolumn{2}{c}{F1 $(\uparrow)$}
          & \multicolumn{2}{c}{AUROC $(\uparrow)$}
          & \multicolumn{2}{c}{AUPRC $(\uparrow)$}\\
          &
          2018 & 2020 &
          2018 & 2020 &
          2018 & 2020 &
          2018 & 2020 &
          2018 & 2020 &
          2018 & 2020 \\
          \midrule
          LightGBM
            & $\mathbf{21.52}$
            & $\mathbf{0.7}$
            & $\mathbf{89.98}$
            & $\mathbf{66.49}$
            & $\mathbf{18.74}$
            & $\mathbf{0.31}$ 
            & $\mathbf{17.15}$
            & $0.48$
            & $\mathbf{93.5}$
            & $\mathbf{75.51}$
            & $\mathbf{34.84}$
            & $\mathbf{2.68}$\\
            & \small{$1.6$}
            & \small{$0.33$}
            & \small{$0.41$}
            & \small{$2.38$}
            & \small{$1.78$}
            & \small{$0.05$}
            & \small{$1.93$}
            & \small{$0.39$}
            & \small{$0.31$}
            & \small{$1.24$}
            & \small{$1.48$}
            & \small{$0.29$}\\
          ECOD
            & $0.48$
            & $0.2$
            & $62.2$
            & $\underline{62.49}$
            & $0.4$
            & $\underline{0.25}$ 
            & $0.57$
            & $1.04$
            & $54.02$
            & $51.59$
            & $1.09$
            & $0.76$\\
            & \small{$0.2$}
            & \small{$0.22$}
            & \small{$0.64$}
            & \small{$1.02$}
            & \small{$0.02$}
            & \small{$0.01$} 
            & \small{$0.26$}
            & \small{$0.38$}
            & \small{$0.54$}
            & \small{$0.87$}
            & \small{$0.06$}
            & \small{$0.04$} \\
          COPOD
            & $0.34$
            & $0.16$
            & $64.77$
            & $\underline{62.64}$
            & $0.43$
            & $\underline{0.25}$ 
            & $0.5$
            & $1.12$
            & $51.7$
            & $50.15$
            & $1.0$
            & $0.73$\\
            & \small{$0.21$}
            & \small{$0.15$}
            & \small{$0.51$}
            & \small{$0.98$}
            & \small{$0.02$}
            & \small{$0.01$} 
            & \small{$0.2$}
            & \small{$0.44$}
            & \small{$0.59$}
            & \small{$0.91$}
            & \small{$0.05$}
            & \small{$0.04$} \\
          Isolation Forest
            & $0.16$
            & $0.19$
            & $64.14$
            & $60.55$
            & $0.43$
            & $0.23$ 
            & $0.71$
            & $\underline{\mathbf{1.3}}$
            & $60.52$
            & $46.86$
            & $1.35$
            & $0.67$\\
            & \small{$0.12$}
            & \small{$0.19$}
            & \small{$0.75$}
            & \small{$0.76$}
            & \small{$0.02$}
            & \small{$0.01$} 
            & \small{$0.23$}
            & \small{$0.34$}
            & \small{$0.52$}
            & \small{$0.84$}
            & \small{$0.07$}
            & \small{$0.03$}\\
          KNN
            & $0.34$
            & $0.01$
            & $\underline{68.87}$
            & $55.6$
            & $0.51$
            & $0.18$ 
            & $0.78$
            & $0.38$
            & $\underline{65.92}$
            & $49.28$
            & $\underline{1.58}$
            & $0.63$\\
            & \small{$0.12$}
            & \small{$0.04$}
            & \small{$0.76$}
            & \small{$0.85$}
            & \small{$0.02$}
            & \small{$0.01$}
            & \small{$0.21$}
            & \small{$0.27$}
            & \small{$0.64$}
            & \small{$0.67$}
            & \small{$0.08$}
            & \small{$0.02$}\\
          GOAD
            & $0.14$
            & $0.19$
            & $53.72$
            & $52.73$
            & $0.17$
            & $0.17$
            & $0.7$
            & $0.69$
            & $50.36$
            & $\underline{64.45}$
            & $0.67$
            & $\underline{1.03}$\\
            & \small{$0.09$}
            & \small{$0.13$}
            & \small{$1.41$}
            & \small{$1.69$}
            & \small{$0.01$}
            & \small{$0.01$}
            & \small{$0.36$}
            & \small{$0.34$}
            & \small{$2.35$}
            & \small{$1.25$}
            & \small{$0.05$}
            & \small{$0.06$} \\
          NeuTraL-AD
            & $0.6$
            & $0.02$
            & $59.12$
            & $51.52$
            & $0.35$
            & $0.15$
            & $1.45$
            & $0.38$
            & $53.23$
            & $45.19$
            & $1.08$
            & $0.58$ \\
            & \small{$0.22$}
            & \small{$0.08$}
            & \small{$3.56$}
            & \small{$1.15$}
            & \small{$0.05$}
            & \small{$0.01$}
            & \small{$0.44$}
            & \small{$0.21$}
            & \small{$1.9$}
            & \small{$1.75$}
            & \small{$0.07$}
            & \small{$0.03$}\\
          Internal Cont.
            & $0.64$
            & $0.0$
            & $39.43$
            & $46.7$
            & $0.18$
            & $0.13$ 
            & $1.21$
            & $0.23$
            & $45.63$
            & $50.66$
            & $0.87$
            & $0.68$\\
            & \small{$0.08$}
            & \small{$0.0$}
            & \small{$1.05$}
            & \small{$0.17$}
            & \small{$0.0$}
            & \small{$0.0$}
            & \small{$0.16$}
            & \small{$0.07$}
            & \small{$2.46$}
            & \small{$0.9$}
            & \small{$0.1$}
            & \small{$0.03$}\\
          NPT-AD 
            & $\underline{0.97}$
            & $\underline{\mathbf{0.66}}$
            & $67.21$
            & $53.2$
            & $\underline{0.81}$
            & $0.18$ 
            & $\underline{1.67}$
            & $0.58$
            & $\underline{66.21}$
            & $53.45$
            & $1.28$
            & $0.61$\\
            & \small{$0.07$}
            & \small{$0.06$}
            & \small{$1.25$}
            & \small{$0.65$}
            & \small{$0.01$}
            & \small{$0.03$}
            & \small{$0.11$}
            & \small{$0.03$}
            & \small{$1.14$}
            & \small{$0.43$}
            & \small{$0.11$}
            & \small{$0.06$}\\
          \bottomrule
          \end{tabular}}
          \label{tab:results}
          \end{table*}
 
        Metrics reported in table \ref{tab:results} are averaged over $10$ runs and we performed t-tests on the highest metrics to asses whether models obtained significantly different results.
        Among the AD approaches, we observe that the non-deep methods demonstrate the best performance, specifically ECOD, COPOD, and KNN. However, it is worth noting that the deep learning approach NPT-AD also yields comparable results. While the AD methods achieve satisfactory results regarding the AUROC, their performances are consistently poor for both the F1-Score and AUPRC metrics across both countries and periods.
        In contrast, LightGBM exhibits significantly better performance across all metrics, consistently achieving the highest values, except for the F1-Score on the 2020 dataset of Country B.
        The overall poor performance of all models, including LightGBM, for the F1-Score and AUPRC, highlights the inherent challenges associated with fraud detection. However, a noteworthy observation is the substantial performance gap between LightGBM and the anomaly detection methods.
            
    \section{Discussion}
    \label{sec:discussion}

    \subsection{Distribution Shift}
    
    The obtained results for both countries support the hypothesis that a distribution shift occurred between 2018 and 2020 in our dataset. Across most tested approaches, we observe a significant decrease in all metrics between the 2018 and 2020 datasets for both countries. Notably, the distribution shift appears more pronounced in country B, with metrics experiencing a more significant decline. The findings presented in Figure \ref{fig:dist_shift} further corroborate this statement as the dataset distance is higher between periods than for country A, and the bi-dimensional representations also show less overlap between the periods than for country A.
    Although the AD methods display poor overall performance, they exhibit more resilience in the face of distribution shift than LightGBM. This trend is particularly evident for ECOD and COPOD, which maintained relatively similar metrics between the 2018 and 2020 datasets for both countries. Conversely, KNN and NPT-AD experienced a significant decline in performance across both periods and datasets compared to ECOD and COPOD. While LightGBM still achieves the highest metrics for most of the 2020 dataset, it suffers a substantial drop in performance between the two periods. This drop is especially pronounced in the AUPRC and F1-Score metrics. Notably, most AD methods outperform LightGBM by a significant margin in terms of the F1-Score for the 2020 dataset of country B.
    
    Based on these findings on our dataset, it appears that LightGBM is a favorable choice in the fraud detection framework when labels are available and no distribution shift occurs. However, in the presence of a distribution shift, retraining LightGBM on an updated dataset becomes crucial to prevent a significant performance decline.

    \subsection{Anomaly Detection Methods for Ensembling}
    
    One potential advantage of anomaly detection (AD) methods is their ability to identify fraud cases that differ from those flagged by supervised approaches. If AD models can successfully detect fraud instances supervised models cannot identify, resorting to ensembling techniques could enhance overall fraud detection performance.
    To investigate this further, we focused on ECOD, one of the top-performing AD methods on the dataset consisting of payments in \textit{country A}. We examined whether the fraud cases detected by ECOD differed from those identified by LightGBM. Across the 10 iterations, we observed that, on average, 3.28\% of the fraud cases in the test set were detected by ECOD but not by LightGBM. Conversely, LightGBM detected 20.41\% of the fraud cases in the test set that ECOD did not flag. Notably, the 3.28\% represents 10.98\% of the fraud cases detected by ECOD. In other words, 89.02\% of the fraud cases detected by ECOD were also detected by LightGBM. As a result, ensembling models by combining AD methods with LightGBM to enhance fraud detection performance may prove ineffective.

    \section{Conclusion}
    \label{sec:conclusion}
    
    In conclusion, our study highlights several key findings concerning applying machine learning techniques for fraud detection. Our results demonstrate that LightGBM consistently outperforms the tested AD methods across various evaluation metrics, emphasizing its efficacy in fraud detection tasks compared to other methods.
    However, we also observed that LightGBM's performances are susceptible to degradation due to distribution shift. This finding underscores the importance of retraining LightGBM on updated datasets when there is suspicion or evidence of a distribution shift. By adapting the model to the changing data distribution, it is possible to mitigate the drop in performance and maintain its effectiveness in fraud detection.
    Furthermore, our investigation revealed that ensembling techniques with AD methods would not significantly improve overall fraud detection performance. Despite the potential for AD methods to detect frauds that may elude supervised approaches, our analysis showed that LightGBM also detected most of the frauds identified by AD methods. This finding suggests limited benefits in combining AD methods with LightGBM in our specific fraud detection framework.
    We believe these insights may contribute to advancing the field of fraud detection and inform practitioners in selecting appropriate models and strategies for robust and accurate fraud detection systems. 
    
    Future work may involve replicating our analysis on other credit card payment datasets to determine whether our obtained results can be generalized. Furthermore, enhancing the robustness of GBDT models against distribution shifts emerges as a critical direction for further exploration. Addressing this challenge is paramount for financial institutions, as it enables them to embrace machine learning techniques in fraud detection systems confidently.

\subsubsection*{Aknowledgements}
This work was granted access to the HPC resources of IDRIS under the allocation 2023-101424 made by GENCI.\newline
This research publication is supported by the Chair "Artificial intelligence applied to credit card fraud detection and automated trading" led by CentraleSupelec and sponsored by the LUSIS company.

\bibliographystyle{styles/bibtex/spmpsci}
\bibliography{140}

\end{document}